\documentclass[a4paper]{article}

\usepackage{INTERSPEECH2018}
\usepackage{enumitem}
\usepackage{graphics}
\usepackage{siunitx}
\usepackage{url}
\usepackage{hyperref}
\newcommand{\repeatthanks}{\textsuperscript{\thefootnote}}

\title{UltraSuite: A Repository of Ultrasound and Acoustic Data\\from Child Speech Therapy Sessions}

\name{Aciel Eshky$^{*}$$^1$\thanks{\hspace*{-.5em}$^{*}$Equal contribution.},
Manuel Sam Ribeiro$^{*}$$^1$\repeatthanks,
Joanne Cleland$^2$,
Korin Richmond$^1$,\\
Zoe Roxburgh$^3$,
James Scobbie$^3$,
Alan Wrench$^{3,4}$}

\address{
  $^1$Centre for Speech Technology Research, University of Edinburgh, UK\\
  $^2$Psychological Sciences and Health, University of Strathclyde, UK\\
  $^3$Clinical Audiology, Speech and Language Research Centre, Queen Margaret University, UK\\
  $^4$Articulate Instruments Ltd., UK
\email{\{aeshky, sam.ribeiro, korin\}@ed.ac.uk 
}}

\begin{document}

\maketitle
\begin{abstract} 
We introduce UltraSuite, a curated repository of ultrasound and acoustic data, collected from recordings of child speech therapy sessions. This release includes three data collections, one from typically developing children and two from children with speech sound disorders. In addition, it includes a set of annotations, some manual and some automatically produced, and software tools to process, transform and visualise the data.  
\end{abstract}


\section{Introduction}
%
Speech sound disorders (SSDs) affect quality of life for a large number of children.
In the UK, 11.4\% of eight year olds\footnote{Native speakers of English are expected to master production of all vowels and consonants by age 8 \cite{wren1prevalence}.} have persistent SSDs, ranging from common clinical distortions to speech that is unintelligible even to close family members \cite{wren1prevalence}. 
SSDs are similarly prevalent in other countries \cite{wren1prevalence}.
Children with disordered speech experience adverse outcomes of many kinds: social and psychological outcomes, difficultly with literacy and educational attainment, and long-term employment prospects \cite{morgan2017refer, Johnson2010Twenty, McCormack2011Nationally, Lewis2011Literacy}.

However, current clinical practice for assessing these disorders is subjective and inaccurate \cite{howard2011cleft}.
Instrumental methods that use articulatory imaging, such as ultrasound, provide a more accurate diagnosis, 
at the expense of large amounts of manual effort from a highly trained pathologist.
Machine learning has the potential to automate much of this work, leading to better outcomes for patients without increasing workload for pathologists,
but publicly available data that could facilitate this work is scarce.
Existing work reports results on adult data \cite{fabre2015tongue, xu2016robust, fabre2017automatic}, 
data that is not publicly available \cite{smith2017improving}, 
or data that is in proprietary format \cite{zharkova2009ultrasound, zharkova2011high, zharkova2016high}. 
%
Additionally, child speech processing and disordered speech processing are both known to present many challenges \cite{russell2007challenges, fainberg2016improving, beckman2017methods, christensen2013combining}. 
Having access to the right kind of data encourages more researchers to work on this problem and compare results.

In this paper we introduce UltraSuite, a curated repository of data obtained from child speech therapy sessions which used articulatory imaging techniques. 
The repository contains synchronised ultrasound and acoustic data recorded with a range of children with different categories of SSDs in addition to typically developing children who were learning new articulations. 
As part of the repository, we release annotations (manually and automatically produced) and software tools to process, transform, and visualise the data. The repository will continue to grow and become larger and more comprehensive as we add new studies, ensuring that all new data is available in the same standardised format. We invite other researchers to contribute their data to  this repository.

\subsection{Motivation}

To better understand the potential for machine learning methods to automate the use of instrumental techniques 
for assessing, diagnosing and treating children with SSDs, it is useful to illustrate how instrumental techniques are used.  


Perception-based methods for assessing SSD are known to be highly subjective \cite{howard2011cleft}. Ultrasound imaging of the tongue provides additional information not available in the acoustic signal (e.g., the presence of double articulations or undifferentiated lingual gestures \cite{gibbon1999undifferentiated}). This additional information reduces subjectivity and in some cases changes the diagnosis.
However, working with speech recordings and ultrasound videos is time consuming and difficult. 

In order to provide a diagnosis or measure therapy progress, the speech pathologist goes through the following process:
searching therapy recordings for occurrences of words of interest 
(the recordings often contain background noise, the voice of the therapist encouraging the child to speak, and the child uttering multiple words and potentially making mistakes); 
identifying boundaries of a phone of interest both in the audio and ultrasound video; 
locating the mid-frame of the phone in the ultrasound video; 
fitting a contour to the tongue in the image; 
and finally, measuring how similar/dissimilar the tongue shape is from the average tongue shape 
in correct articulation. 
While this process offers a more accurate diagnosis it is time consuming, tedious, and requires specialist training and is therefore not offered clinically. Automating it would allow it to be offered to children as standard practice.

\subsection{Broader Applicability of the Data}
Access to the articulatory domain through imaging techniques such as ultrasound gives additional information over the acoustic domain. Indeed, in addition to the area of speech and language pathology, prior work has shown that articulatory information has the potential to improve performance in multiple aspects of speech technology, for instance: speech recognition \cite{king2007speech}, speech synthesis \cite{richmond2015use}, and silent speech interfaces \cite{denby2010silent}.


\subsection{Paper Outline}
In Sections \ref{sec:collection} and \ref{sec:prep} we describe the process of collecting and standardising the data and include a description of a train-test split for a subset of the data. 
In Section \ref{sec:SD} we describe our annotation work, followed by data statistics in Section \ref{sec:stats}, a description of the software tools in Section \ref{sec:tools}, and license information in section \ref{sec:license}. We conclude in Section \ref{sec:future} with a brief description of data we are currently in the process of collecting with the aim of including in the repository in the future. 

\begin{table}[t]
  \caption{The number of participants, their gender and ages. We report ages in years (y) and months (m). We recorded the ages of participants on their single visit in UXTD, and on their first baseline in UXSSD and UPX.}
  \label{tab:ages}
  \centering
  \begin{tabular}{llll}
    \toprule
    & \textbf{UXTD} & \textbf{UXSSD} & \textbf{UPX} \\
    \midrule
    Number of participants & 58 & 8 & 20 \\
    Female & 31 & 2 & 4 \\
    Male & 27 & 6 & 16 \\
    Mean age & 9y 3m & 7y 7m & 8y 4m \\
    SD age & 1y 10m & 1y 6m & 2y 2m \\    
    Min age & 5y 8m & 5y 11m & 6y 1m \\
    Max age & 12y 10m & 10y 1m & 13y 4m \\
    \bottomrule
  \end{tabular}
\end{table}

\section{Data Collection}\label{sec:collection}
Ethical approval to collect the data was granted by the NHS Research Ethics Service. We recorded the data in the laboratory using the Articulate Assistant Advanced software (AAA), initially storing it in the AAA proprietary data format \cite{articulate2010articulate}.
All sessions were conducted by a speech and language therapist (SLT), and both the children and the therapists spoke English with a standard Scottish accent. All therapists were female. 
We collected three datasets, one with typically developing children (TD) and two with children with speech sound disorders (SSD). 
The participants' guardians provided consent to allow the data to be made available to the research community\footnote{We excluded from the repository participants whose guardians did not provide consent.}. 

A set of prompts specified the verbal task the child was expected to perform; for example, sentences to be read, isolated phones to be uttered, or pictures to be described. 
For each utterance, we recorded the acoustic signal and an ultrasound video of the child's mouth. We placed the ultrasound transducer probe submentally (under the chin) capturing the midsagittal view of the child's tongue \cite{stone2005guide} and stabilised it using a headset.
Often the child needed encouragement to speak, so the acoustic signal contains both the SLT's speech and the child's speech. The child didn't always stick to the prompt; they hesitated, repeated, or made mistakes. In picture-describing tasks the speech is conversational in nature (e.g., ``what's this?" ``a frog stuck in a spiderweb" ``aha, anything else?" ``a strawberry in his mouth"). The prompt is therefore not a transcription of the audio.

\subsection{Typically Developing Children}
We recorded the typically developing subset of the Ultrax dataset (UXTD) between 11/2011--10/2012. The purpose of the experiment was to evaluate the effectiveness of ultrasound as a visual biofeedback tool for learning non-English articulations \cite{cleland2015helping}. Each child attended once and recorded a single session.  

\subsection{Children with Speech Sound Disorders}
The repository at this stage contains two datasets recorded with children with speech sound disorders. The first is the Ultrax speech sound disorders subset (UXSSD) which we recorded between 12/2011--07/2014, and the second is the UltraPhonix dataset (UPX), recorded between 06/2015--03/2017. The children exhibited a range of SSDs including phonological delay, phonological disorder, inconsistent phonological disorder, vowel disorder, articulation disorder, and childhood apraxia of speech. The data was recorded specifically for the purpose of evaluating the effectiveness of ultrasound as a visual biofeedback tool for therapy \cite{cleland2015using, cleland2017covert}. 
Each child attended several sessions: suitability (before baseline), baseline (1--5 sessions), therapy (1--12 sessions), mid-therapy, post-therapy (immediately after therapy), and maintenance (several months after therapy). Table~\ref{tab:ages} shows the number of participants in each of the three datasets, their gender and ages. Persistent SSDs are more commonly associated with boys than girls \cite{wren1prevalence}, which explains the gender imbalance in the data.
\section{Data Preparation}\label{sec:prep}
We exported the raw data from the proprietary AAA format to obtain a tuple of four files per utterance: 
\begin{enumerate}
\item \textbf{Prompt file}: contains text describing the task the child was given and the date-time of recording.
\item \textbf{Audio file}: RIFF wave file, sampled at 22.05 KHz, containing the speech of the child and the SLT. 
\item \textbf{Ultrasound file}: a sequence of ultrasound frames capturing the midsagittal view of the child's tongue. 
A single ultrasound frame is recorded as a 2D matrix where each column represents the ultrasound reflection intensities along a single scanline. The surface of the probe is convex and the scanlines are directed in an equal-angled fan in the scanning plane. In order to correctly interpret the ultrasound data, a set of parameters are recorded in the parameter file described below.  
\item \textbf{Parameter file}: contains a set of parameters to interpret the ultrasound data and synchronise it with the audio. It gives the number of scanlines in each frame (63), the number of data points per scanline (412), number of bits used to represent each reflection intensity data point (8), the angle between each scanline (0.038$^{\circ}$), the number of ultrasound frames per second ($\approx$121.5 fps), and a synchronisation offset relative to the audio in seconds. 
\end{enumerate}
We discarded utterance tuples where the audio was too short and was unrelated to the prompt.

\subsection{Prompts}
\begin{table}[t]
  \caption{The number of utterances per prompt type, with the number of unique prompts in parentheses for each of the three datasets. We encode the type identifier in the file names.} 
  \label{tab:prompt_type}
  \centering
  \tabcolsep=0.18cm
  \begin{tabular}{lllll}
    \toprule
    \textbf{Type} & \textbf{ID} & \textbf{UXTD} & \textbf{UXSSD} & \textbf{UPX}   \\
    \midrule
    Words & A & 962 (26) & 2708 (291) & 3838 (455)\\
    Non-words & B & 607 (27) & 495 (59) & 560 (60)\\
    Sentence & C & 0 (0) & 445 (35)	& 1020 (128)\\
    Articulatory & D & 2934 (45) & 132 (17) & 211 (31)\\
    Non-speech & E & 116 (2) & 9 (1) & 302 (1)\\
    Other & F & 0 (0) & 56 (12) & 61 (7)\\
    \midrule
    \multicolumn{2}{l}{Total}   & 4619 (100) & 3845 (415)  & 5992 (682)  \\
    \bottomrule
  \end{tabular}
\end{table}
We standardised the formatting of the prompt text by removing inconsistencies, such as replacing tabs with white spaces, removing duplicate or trailing white spaces, correcting the capitalisation of proper nouns, and correcting misspellings. We identified six distinct types of prompts:
\begin{enumerate}[label=(\Alph*)]
\item \textbf{Words}: a group of semantically unrelated English words (e.g., ``down link pat get") which were either identified by the SLT as being diagnostically useful, or were based on a protocol from the Diagnostic Evaluation of Articulation and Phonology (DEAP) \cite{dodd2002diagnostic}. 
\item \textbf{Non-words}: designed to elicit certain phones from the child but which are not real words (e.g., ``p apa epe opo").
\item \textbf{Sentence}: designed to elicit co-articulation (e.g., ``It's a toe Pam") or designed to examine phones of interest in different contexts at the sentence level (e.g.,``My Granny Maggie got a golden gown" where /g/ occurs in different word positions and different vowel environments). 
\item \textbf{Articulatory}: single or multiple phones occurring once or repeated. The SLT pronounces a phone, or plays a recording of a phone at different speeds, and the child is expected to imitate what they hear. The latter is known as a Diadochokinesis imitation task \cite{mccann2007new}.
\item \textbf{Non-speech}: includes swallowing motions recorded to obtain a trace of the hard palate, and coughs recorded to obtain additional tongue shapes.
\item \textbf{Other}: conversational speech, such as describing a picture or telling a story about it (e.g., ``Connected speech picture 1").
\end{enumerate}
The number of utterances per prompt type in each dataset and the number of unique prompts are shown in Table \ref{tab:prompt_type}.

\subsection{File Naming Convention} \label{sec:naming}
We placed each session in a directory and labelled it accordingly (Suit, BL, Therapy, Mid, Post, and Maint). Typically developing children recorded a single session each, so the UXTD directories are labelled with speaker identifiers only.
Within each session, we sorted the utterances by the date-time of recording and indexed them from 001. We then appended the prompt type identifier (A-F) to the index. For example, if the 5th utterance in a session is a sentence, the tuple of files associated with the utterances are 005C.txt, 005C.wav, 005C.ult, and 005C.param.

\subsection{Train, Development, and Test Splits}
We split the UXTD dataset into training, development, and testing subsets balancing by gender and age.
Training contains 40 children (18 male, 22 female), development 6 children (3 male, 3 female), and testing 12 children (6 male, 6 female).
The mean and standard deviation of the children's age in each subset is:
9y 5m $\pm$ 1y 10m, 
9y 1m $\pm$ 1y 9m, and 
8y 12m $\pm$ 1y 10m.

We omit a split for UXSSD and UPX due to the small number of participants in each SSD subcategory. However, we urge users of this data to report the ID of the participants and the name of sessions used for training and testing. 

\section{Data Annotation}\label{sec:SD} 
In addition to the data described in the previous section, we release a set of annotations, including pronunciation dictionaries for each of the datasets, audio transcriptions for UXTD, SLT annotations, automatic speaker labelling and automatic phone alignments, all of which can aid modelling. 


\textbf{Pronunciation dictionaries:}
We prepared a pronunciation dictionary for each of the three datasets.
We did this by listing the words that appear in the prompts, looking them up in a standard lexicon, and copying their phonetic transcription. We used a Scottish accent variant of the Combilex lexicon 
to match the accent in the data \cite{richmond2009robust, richmond2010generating}.
For out-of-vocabulary words, such as the non-words of type B prompts, an annotator with training in phonetics transcribed their expected pronunciation. 
The vocabulary size is 296, 1048, and 1437 for the UXTD, UXSSD, and UPX datasets respectively. 

\textbf{SLT labelling:} 
The SLTs annotated a small portion of the data for the intervention studies the data was originally collected for \cite{cleland2015helping, cleland2015using, cleland2017covert}. 
The annotations include boundaries of words and phones of interest, and tongue contours manually fitted to the mid-frame of phones of interest in the ultrasound video.
The number of utterances with at least one label are 3900, 152, and 3919 in the UXTD, UXSSD, and UPX datasets, respectively. 
We release these annotation as Praat's TextGrid files \cite{Boersma2009} and follow the same naming convention described in Section \ref{sec:naming}.

\textbf{Audio Transcriptions:} 
Because the audio recording is not a direct match to the prompt, we provide a small subset of transcribed utterances, namely utterances of types A and B for all speakers in the UXTD dataset.
A single annotator listened to the audio and transcribed the child's speech.
SLT intervention is loosely transcribed as \emph{[SLT:token]}, where \emph{token} takes the form of \emph{spn} (spoken sound) to denote generic SLT speech, or the form of a word if that word occurs in the prompt, for example \emph{[SLT:helicopter]}.
It is less obvious how to transcribe disordered speech, we therefore do not provide transcriptions for the UXSSD and UPX datasets.

\textbf{Speaker labelling:} 
In order to attribute different parts of the audio to different speakers, and to quantify the hours of speech, 
we trained a model that discriminates between SLT and child speech. 
We used the transcriptions of the UXTD dataset as training data by reducing words to \emph{child} and \emph{SLT} tokens, corresponding to turn-taking sequences between therapist and participant.
Using Kaldi's \cite{povey2011kaldi} standard monophone recipe, we modelled these tokens with 5-state ergodic HMMs \cite{matsui1994comparison}.
Silences were modelled with 5 state left-to-right skip HMMs.
As a post-processing step, we merged identical labels that were separated by a silence with less than 100ms.
A second pass of the data then removed labels with duration less than 50ms.

To measure the accuracy of this method, we used the force-aligned transcriptions of the UXTD's test set as a ground truth.
We estimated error using \emph{pyannote.metrics} \cite{bredin2017pyannote}, computing error in terms of seconds.
We observed an Identification Error Rate of 4.6\%, and precision and recall of 0.969 and 0.979, respectively.
We decoded the three datasets with this method, which forms the basis for the data reported in Table \ref{tab:dataset_stats}.


\textbf{Phone labelling:} 
Automatically identifying phones in a child's speech would significantly reduce the workload for an SLT. 
As an initial solution, we applied standard phone alignment to the data. To obtain additional training data, we pooled the training subset of UXTD and the PF-STAR corpus \cite{batliner2005pf_star}, and trained a phone alignment model following the PF-STAR baseline recipe presented in \cite{fainberg2016improving}.

Using our pronunciation dictionaries, we substituted the words in the prompts and audio transcriptions with their phonetic transcriptions for utterances of types A, B, and C.
We then aligned the waveforms to the transcriptions in UXTD, and to the prompts in UXSSD and UPX since transcriptions are not available for these two datasets. Disordered speech is therefore aligned to expected pronunciation rather than true pronunciation.  

Although we used standard methods to obtain phone alignments, these datasets pose significant challenges.
Besides well known difficulties associated with child speech processing \cite{russell2007challenges} and disordered speech processing \cite{christensen2013combining}, additional issues include variability in the recording conditions, interaction between the therapist and child, and deviations from the prompts.
Future work will investigate more robust methods for phone alignment and ways of evaluating them.


\section{Data Statistics}\label{sec:stats}
\begin{table}[t]
  \caption{
  Hours of speech and silence rounded to two decimal places, estimated using the speaker labelling method described in Section \ref{sec:SD}, with percentages given in parentheses.}
  \label{tab:dataset_stats}
  \centering
  \resizebox{\columnwidth}{!}{%
  \begin{tabular}{lllll}
    \toprule
     & \textbf{UXTD} & \textbf{UXSSD} & \textbf{UPX} \\
    \midrule
    Child speech & 2.24 (28.39\%)           & 3.66 (34.45\%)          & 7.27 (38.70\%) \\ 
    SLT speech   & 1.24 (15.74\%)           & 1.81 (16.99\%)          & 1.92 (10.23\%) \\
    \textbf{Total speech} & 3.47 (44.12\%)  & 5.47 (51.43\%)          & 9.19 (48.93\%) \\
    \midrule
    Initial silence & 1.41 (17.96\%)        & 0.91 (8.55\%)           & 0.78 (4.17\%) \\
    Medial  silence & 1.99 (25.30\%)        & 3.48 (32.69\%)          & 7.11 (37.83\%) \\
    Final   silence & 0.99 (12.61\%)        & 0.78 (7.32\%)           & 1.70 (9.07\%) \\
    \textbf{Total silence} & 4.40 (55.88\%) & 5.16 (48.57\%)          & 9.59 (51.07\%) \\
    \midrule
    \textbf{Total audio} & 7.87 & 10.63 & 18.78 \\
    \bottomrule
  \end{tabular}
  }
\end{table}

Overall, the data contains 37.28 hours of synchronised audio and raw ultrasound across all datasets.
Table \ref{tab:dataset_stats} shows the distribution of audio in terms of speech (child and SLT) and silences (utterance initial, medial, and final), estimated using the speaker labelling method described in Section \ref{sec:SD}.
Although our speaker labelling method achieved good results on UXTD's test set, it was not directly evaluated on articulatory tasks or disordered speech. An inspection of the assigned labels in the data showed missed cases of child speech, especially when decoding utterances of type D. The estimates of speech shown in Table \ref{tab:dataset_stats} are therefore conservative.

We estimated a total of 18.67 hours of speech in the three datasets. Despite this being a conservative estimate, it is comparable to the number of hours of speech in the standard child speech corpus PF-STAR, 
which has 10 hours of read speech by native English speaking children aged 6-11, 
and 10 additional hours of spontaneous speech by native English speaking children aged 4-14 
\cite{batliner2005pf_star}.
We preserve initial and final silences in the data as the corresponding ultrasound may be useful for other tasks, such as tongue contour extraction.
We estimated 91.83, 14.02, and 28.25 minutes of child speech for the training, development, and testing subsets of the UXTD dataset.

\section{Companion Code Repository}\label{sec:tools}
We distribute a code repository containing a set of tools to interpret, transform and visualise the data, in addition to the recipes used to annotate the data. We describe the current contents of the code repository and invite users to contribute their own code. 

\begin{figure}[t]
  \centering
  \includegraphics[width=\linewidth]{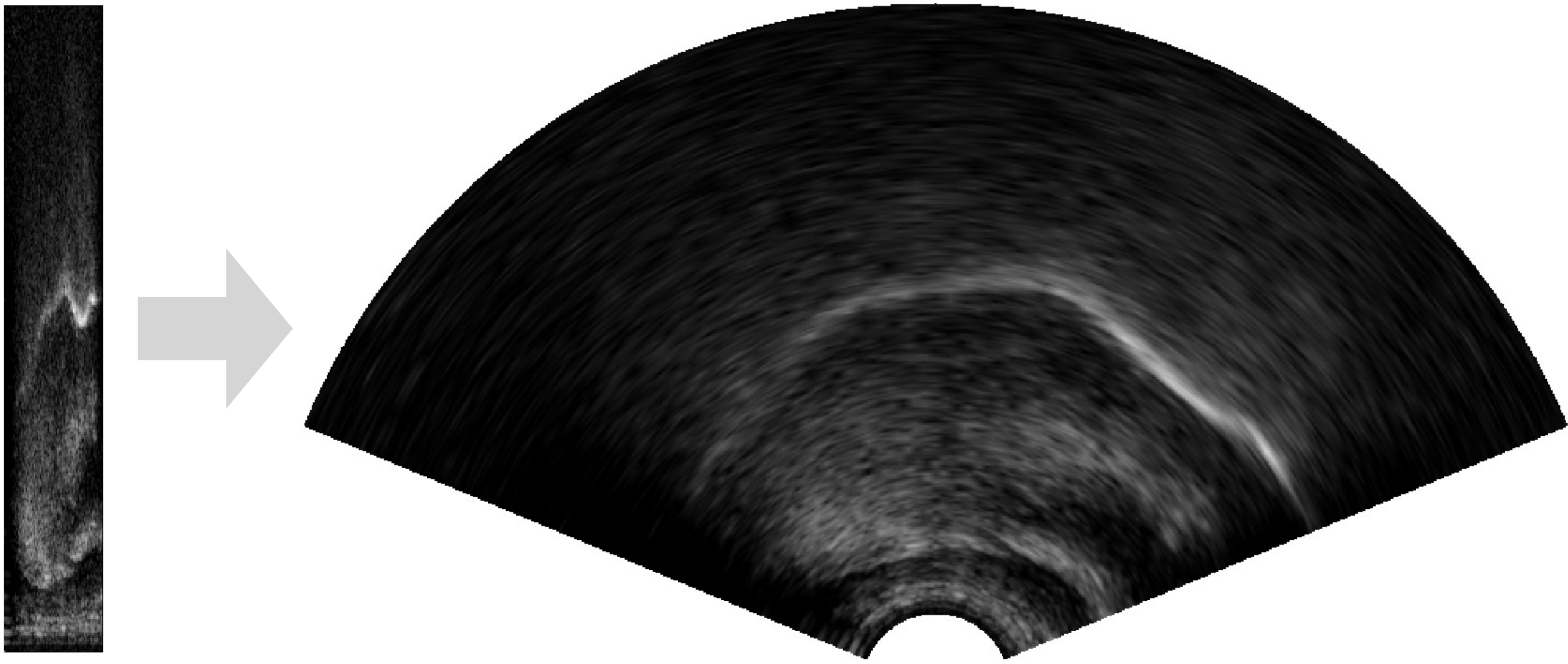}
  \caption{An ultrasound image showing the midsagittal view of a child's tongue. We store the raw ultrasound reflection data efficiently as a matrix (left), but provide a tool to transform it to real world proportions (right).}
  \label{fig:ultrasound}
\end{figure}

\textbf{Tools:} The repository contains raw ultrasound reflection data, but we provide a set of tools to transform it for visualisation. A raw ultrasound file is a sequence of 2D matrices (or a 3D array) where each matrix is a frame, and each column in a frame contains ultrasound reflection data of a single scanline. To correctly interpret the ultrasound data, we provide a tool to transform the raw representation to the real world proportions. The function interpolates the spaces between the scanlines and the result is visualised as a fan image. Figure \ref{fig:ultrasound} illustrates this process. Another tool produces a sequence of images or a video from a raw ultrasound file. 

\textbf{Recipes:} We release the Kaldi recipes 
which we used to train the speaker and phone labelling models.

\section{License and Distribution}\label{sec:license}
We distribute UltraSuite under Attribution-NonCommercial 4.0 Generic 
\begin{footnotesize}(CC BY-NC 4.0)\end{footnotesize}
and distribute the companion code under Apache License v.2.  
Both can be obtained from the project website: 
\begin{footnotesize}\url{http://www.ultrax-speech.org/ultrasuite}\end{footnotesize}

\section{Conclusions and Future Work}\label{sec:future}
We have introduced a new repository of ultrasound and acoustic data which we have collected from child speech therapy sessions. We have described the process of data collection, preparation and standardisation, along with a suggested train-test split. We have described tools to transform and visualise the data, and annotations including pronunciation dictionaries, audio transcriptions, SLT annotations, automatic speaker labelling and automatic phone alignments.   

We will continue to grow the repository by adding more data and tools.
We are in the process of collecting further data from 120 children with SSD in the Ultrax2020 project following the protocol described in \cite{cleland2018ultrax2020}.
In addition, we intend to add other available data to our repository, including adult data and alternative forms of articulatory imaging techniques (e.g., MRI of vocal tracts), all of which can be used in data augmentation methods \cite{christensen2013combining, fainberg2016improving, smith2017improving}.
We encourage other researchers to contribute by submitting their data for us to standardise and add to this repository.

\section{Acknowledgements}
Supported by: EPSRC Healthcare Partnerships Programme, grants number
\begin{footnotesize}EP/I027696/1\end{footnotesize} (Ultrax) and \begin{footnotesize}EP/P02338X/1\end{footnotesize} (Ultrax2020), and 
NHS Scotland CSO, grant number \begin{footnotesize}ETM/402\end{footnotesize} (UltraPhonix). 
We thank 
Steve Renals for continued guidance and support,
Anna Womack for help collecting the UltraPhonix data and Steve Cowen for technical support. We thank the children for participating and their guardians for providing consent.

\bibliographystyle{IEEEtran}

\bibliography{dataset_paper_biblio}

\end{document}